\documentclass{article} 
\usepackage{iclr2023_conference,times}

\usepackage{amsmath,amsfonts,bm}

\def\eqref#1{equation~\ref{#1}}

\def\1{\bm{1}}

\def\vc{{\bm{c}}}

\def\vv{{\bm{v}}}

\def\vx{{\bm{x}}}

\DeclareMathAlphabet{\mathsfit}{\encodingdefault}{\sfdefault}{m}{sl}
\SetMathAlphabet{\mathsfit}{bold}{\encodingdefault}{\sfdefault}{bx}{n}

\usepackage{hyperref}
\usepackage{url}
\usepackage{wrapfig}
\usepackage{booktabs}       
\usepackage{amsfonts}       
\usepackage{nicefrac}       
\usepackage{microtype}      
\usepackage{xcolor}         
\usepackage[utf8]{inputenc} 
\usepackage[T1]{fontenc}    

\usepackage{amsmath}
\usepackage{comment}
\usepackage{multirow}
\usepackage{threeparttable}
\usepackage{graphicx}
\usepackage{float} 
\usepackage{subfigure} 
\usepackage{makecell}

\usepackage{amssymb}
\usepackage{amsmath}
\usepackage{pifont}
\newcommand{\xmark}{\ding{55}}%
\usepackage{caption}
\newcommand{\han}[1]{\textcolor{purple}{#1}}

\title{Transferable Unlearnable Examples}

\author{Jie Ren\thanks{Equal contribution.} \\
Michigan State University\\
\texttt{renjie3@msu.edu} \\
\And
Han Xu$^\ast$ \\
Michigan State University\\
\texttt{xuhan1@msu.edu} \\
\And
Yuxuan Wan \\
Michigan State University\\
\texttt{wanyuxua@msu.edu} \\
\And
Xingjun Ma \\
Fudan University\\
\texttt{xingjunma@fudan.edu.cn} \\
\And
Lichao Sun \\
Lehigh University\\
\texttt{lis221@lehigh.edu} \\
\And
Jiliang Tang \\
Michigan State University\\
\texttt{tangjili@msu.edu} \\
}

\begin{document}

\maketitle

\begin{abstract}
With more people publishing their personal data online, unauthorized data usage has become a serious concern. The \textit{unlearnable} strategies have been introduced to prevent third parties from training on the data without permission. They add perturbations to the users' data before publishing, which aims to make the models trained on the perturbed published dataset invalidated. These perturbations have been generated for a specific training setting and a target dataset. However, their unlearnable effects significantly decrease when used in other training settings and datasets. 
To tackle this issue, 
we propose a novel unlearnable strategy based on \textit{Classwise Separability Discriminant} (CSD), which aims to better transfer the unlearnable effects to other training settings and datasets by enhancing the linear separability.
Extensive experiments demonstrate the transferability of the proposed unlearnable examples across training settings and datasets. 
\end{abstract}

\section{Introduction}
\label{sec:intro}









\vspace{-0.1in}

With more people publishing their personal data online, it has raised the concern that the published data can be utilized without the data owner's permission to train machine learning models unauthorizedly. Large-scale datasets collected from Internet like LFW\citep{huang2008labeled}, Freebase\citep{bollacker2008freebase}, and Ms-celeb-1m\citep{guo2016ms} promote the development of deep learning. However, they also have the potential risk of privacy leakage.
Thus, growing efforts~\citep{huang2020unlearnable, fowl2021adversarial} have been made on protecting data from unauthorized usage 
by making the data samples unlearnable~\citep{huang2020unlearnable,fowl2021adversarial,he2022indiscriminate}. In these methods, they generate the unlearnable examples by injecting imperceptible ``shortcut'' perturbation. If the data is used by unauthorized training, the models prefer to extract such easy-to-learn shortcut features and ignore the semantic information in the original data~\citep{geirhos2020shortcut}. Thus, the trained model fails to recognize the user's data during the test phase and the user's data gets protected.

However, the majority of existing methods have weaknesses in two types of transferability, training-wise and data-wise, which can largely limit the practical use of unlearnable examples. \textbf{First,} weak training-wise transferability implies that the perturbed samples generated towards one target training setting, such as ERM, cannot be transferred to other training settings. As shown in Section~\ref{sec:pre_training_wise_transfer},  although Error-Minimizing Noise (EMN) ~\citep{huang2020unlearnable} can protect data from supervised training, we can use unsupervised learning methods, such as Contrastive Learning~\citep{chen2020simple, chen2021exploring, chen2020improved}, to first learn useful representations from the EMN-perturbed dataset, and then fine-tune the model on the perturbed dataset. In this way, the unsupervised model can also achieve comparable performance with the model that is trained on the unperturbed dataset. 
\textbf{Second}, insufficient data-wise transferability indicates that the unlearnable effect of perturbations generated for one target dataset will significantly decrease when transferred to other datasets. Without effective data-wise transferability, we have to generate perturbations for each dataset, which makes the unlearnable process inflexible in reality. For example, data in various applications such as social media is dynamic or even streaming and it is challenging to generate the entire perturbation set when new data is continuously emerging. 



In this work, we aim to enhance the training-wise and data-wise transferability of unlearnable examples. In detail, our method is motivated by the method {Synthetic Noise (SN)}~\citep{yu2021indiscriminate}, which devises a manually designed linear separable perturbation to generate unlearnable examples. 
Such perturbation does not target specific dataset, thus it has the potential to enhance  data-wise transferability. However, SN is manually designed and it is not quantifiable or optimizable. 
Therefore, it is impossible to incorporate SN into other optimization processes.
Meanwhile, SN lacks training-wise transferability. 
Therefore, in our paper, we propose \textit{Classwise Separability Discriminant} (CSD) to generate optimizable linear-separable perturbations. Our framework \textit{Transferable Unlearnable Examples} with enhanced linear separability can generate unlearnable examples with superior training-wise and data-wise transferability.
\section{Related Work}
\label{sec:relared}
\vspace{-0.12in}

\noindent\textbf{Unlearnable Examples.} Unlearnable examples are close to availability attack which aims at making the data out of service for training the models by the third parties ~\citep{munoz2017towards}. Before releasing the data in public, we can make it unlearnable to stop others from using it for training ML model without permission. Several works produce unlearnable examples with the guidance of the label information ~\citep{huang2020unlearnable, fowl2021adversarial, shan2020fawkes,yu2021indiscriminate}. The vanilla unlearnable examples are produced by an alternating bi-level min-min optimization on both model parameters and perturbations~\citep{huang2020unlearnable}. Being induced to trust that the perturbation can minimize the loss better than the original image features, the model will pay more attention to the perturbations. ~\cite{yu2021indiscriminate} pointed out that a common property behind the perturbations is linear separability in input space.
We refer to unlearnable examples generated under the supervised setting as {\bf supervised unlearnable examples}. Very recently,  Unlearnable Contrastive Learning (UCL) is proposed~\cite{he2022indiscriminate} to generate unlearnable examples based on unsupervised contrastive learning and to protect data from unsupervised learning. We refer to unlearnable examples generated under the unsupervised setting as {\bf unsupervised unlearnable examples}.  Our studies show that both supervised and unsupervised unlearnable examples lack training-wise transferability as shown in Section \ref{sec:pre_training_wise_transfer} and Appendix \ref{append_unsupervised}, respectively.   Table~\ref{tab:setting} summarizes the settings of existing methods, where \xmark \ means ineffective protection. 
\begin{table}[h]
\vspace{-0.1in}
  \caption{Unlearnable Examples Settings of Existing Approaches}
  \vspace{-0.13in}
  \label{tab:setting}
  \centering
  \begin{tabular}{c|cc}
    \toprule
    Prevented unauthorized use & Supervised Training & Unsupervised Training \\
    \midrule
    Supervised Unlearnability & \makecell{\cite{huang2020unlearnable, fowl2021adversarial}} & \xmark  \\
    \midrule
    Unsupervised Unlearnability & \xmark & \cite{he2022indiscriminate} \\
    \bottomrule
  \end{tabular}
\end{table}
\vspace{-0.2in}

\noindent\textbf{Unsupervised Learning.}
Recently, unsupervised learning has shown its great potential to learn the representation from unlabeled data. Contrastive learning, one of the popular unsupervised methods in computer vision, uses the task of instance discrimination to learn the representations. In SimCLR~\citep{he2022indiscriminate} which is the most common contrastive learning method, the positive and negative samples for each instance are created and the task is to discriminate the positive samples and negative samples. Some improved methods like SimSiam~\citep{chen2021exploring} and BYOL~\citep{grill2020bootstrap} can remove the negative samples and change the task to 
pushing the representations between positive samples to be similar. Many works have been proposed to explain why instance discrimination can lead to good representations. It is claimed in~\cite{wang2020understanding} that uniformity and alignment in feature space are the keys to a good representation. 


\section{Preliminary}
\label{sec:prelimi}

\vspace{-0.1in}




In this section, preliminary studies are conducted to explore two types of transferability.  We first introduce key notations and definitions, and then show the insufficiency of transferability. {\bf Since the majority of unlearnable examples are generated under the supervised setting, in this work, we focus on supervised unlearnable examples and leave unsupervised unlearnable examples as one future work.}  Note that the observations of supervised unlearnable examples in terms of transferability could be applicable to unsupervised unlearnable examples. For example in Appendix \ref{append_unsupervised}, we show that unsupervised unlearnable examples also lack training-wise transferability. 

\vspace{-0.05in}

\subsection{{Definitions}}
\vspace{-0.1in}

In this subsection, we first give the definition of unlearnable examples and then define the training-wise and data-wise transferability.
\vspace{-0.05in}

\textbf{Unlearnable Examples.}
Suppose that the training dataset contains $n$ clean examples $\mathcal{D}_c=\{(\vx_i, y_i)\}_{i=1}^n$ with the input data $\vx_i \in \mathcal{X} \subset \mathbb{R}^d$ and the associated label $y_i \in \mathcal{Y} = \{1, 2, \dots, K\}$.
We assume that the unauthorized parties will use the published training dataset to train a classifier $f_{\theta}: \mathcal{X} \rightarrow \mathcal{Y}$, where $\theta$ is the model parameters, and do inference on their test dataset. To protect the data from unauthorized training, instead of publishing $\mathcal{D}_c$, we want to generate an unlearnable dataset as $\mathcal{D}_u=\{(\vx_i + \boldsymbol{\delta}_i, y_i)\}_{i=1}^n$ where $\boldsymbol{\delta}_i \in \Delta_{\mathcal{D}_c} \subset \mathbb{R}^d$ and $\Delta_{\mathcal{D}_c}$ is the perturbation set for $\mathcal{D}_c$. 
The goal of unlearnability is that if we only publish $\mathcal{D}_u$, the model $f_{\theta}$ trained on $\mathcal{D}_u$ performs poorly on the test dataset. Benefiting from the constraint $\|\boldsymbol{\delta}\|_{p} \leq \epsilon$, $\mathcal{D}_u$ appears the same as $\mathcal{D}_c$ in human eyes and does not affect the normal use.
The most representative EMN~\cite{huang2020unlearnable} generates supervised unlearnable examples by alternating optimization on the bi-level min-min problem:
\vspace{-0.07in}
\begin{align}
\label{EMN_formula}
    \min _{\theta} \min _{\boldsymbol{\delta}_{i} \in \left\{\vv: \left\|\vv\right\|_{\infty} \leq \epsilon\right\}} \sum_{i=1}^{n} \mathcal{L}\left(f_{\theta}\left(\mathbf{x}_{i}+\boldsymbol{\delta}_{i}\right), y_i\right).
\end{align}
\vspace{-0.2in}

The outer minimization can imitate the training process, while the inner minimization can induce $\boldsymbol{\delta}_i$ to have the property of minimizing the supervised loss. Due to this property, deep models will pay more attention to the easy-to-learn $\boldsymbol{\delta}_i$ and ignore $\vx_i$. 

\vspace{-0.05in}

\textbf{Training-wise Transferability.} Supervised unlearnable examples have been designed to protect data from supervised training~\cite{huang2020unlearnable}. However, the unlearnable effect is almost lost when they are utilized for unsupervised training. The unauthorized parties can first get a useful feature extractor $g_\eta$ from $\mathcal{D}_u$ by unsupervised training like Contrastive Learning~\citep{chen2020simple} and then fine-tunes on $\mathcal{D}_u$ or other data to get the classifier $h_g$. The training-wise transferability means that supervised unlearnable examples can invalidate $g_\eta$ when transferred into unsupervised training.

\vspace{-0.05in}

\textbf{Data-wise Transferability.}
When any other dataset, $\mathcal{D}_{\tilde{c}} = \{(\tilde{\vx}_i, \tilde{y}_i)\}_{i=1}^{\tilde{n}}$, where $\tilde{\vx}_i \in \tilde{\mathcal{X}} \subset \mathbb{R}^d$ and $\tilde{y}_i \in \tilde{\mathcal{Y}} = \{1, 2, \dots, \tilde{K}\}$, also requires protection, it is more efficient and practical if we can transfer the perturbation that has already been generated for $\mathcal{D}_{c}$ onto $\mathcal{D}_{\tilde{c}}$. If the perturbation, $\Delta_{\mathcal{D}_c}$, is data-wise transferable, we can also create an unlearnable dataset $\mathcal{D}_{\tilde{u}}$ to replace $\mathcal{D}_{\tilde{c}}$ before publishing as 
$
    \mathcal{D}_{\tilde{u}}=\{(\tilde{\vx}_i + \boldsymbol{\delta}_{H(i)}, \tilde{y}_i)\}_{i=1}^n,
$
where $~\tilde{\vx}_i\in \mathcal{D}_{\tilde{c}}$, $\boldsymbol{\delta}_{H(i)} \in \Delta_{\mathcal{D}_c}$, and $H(i)$ decides which perturbation in $\Delta_{\mathcal{D}_c}$ is added on $~\tilde{\vx}_i$ in the new dataset, $\mathcal{D}_{\tilde{c}}$. 
Without retraining the perturbation set, the unseen $\mathcal{D}_{\tilde{c}}$ is also protected by transferring the unlearnable perturbations from $\Delta_{\mathcal{D}_c}$.

\vspace{-0.01in}
\subsection{Transferability in Existing Methods}
\label{sec:transfer_exitsing}
\vspace{-0.05in}

In this subsection, we show that existing supervised unlearnable examples have almost no training-wise transferability and insufficient data-wise transferability.
\vspace{-0.05in}

\subsubsection{Training-wise Transferability}
\label{sec:pre_training_wise_transfer}

\begin{wrapfigure}{r}{0.32\textwidth}
\begin{center}
\vspace{-0.8in}
\includegraphics[width=0.3\textwidth]{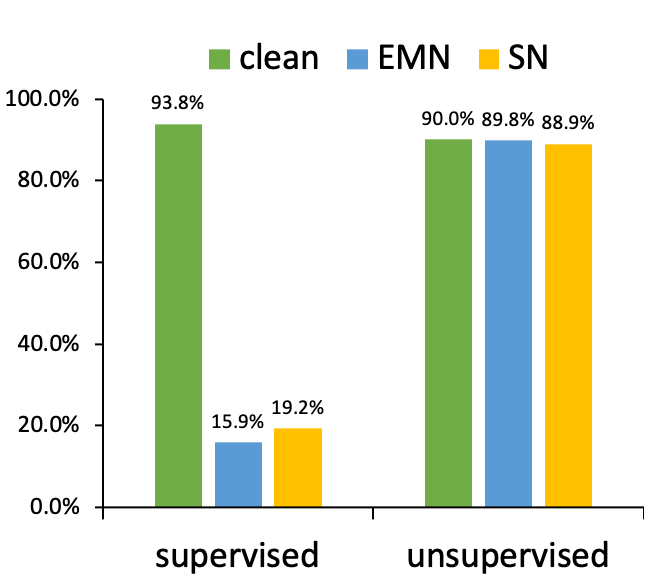}
\end{center}
\vspace{-0.2in}
\caption{Accuracy on clean test data of supervised and unsupervised models on trained on clean CIFAR-10 and EMN-pertubed CIFAR-10}
\vspace{-0.2in}
\label{intro_pic}
\end{wrapfigure}

\vspace{-0.05in}

In Figure~\ref{intro_pic}, experiments are provided to show the ineffective protection of two supervised unlearnable examples, i.e. Error-Minimizing Noise (EMN)~\citep{huang2020unlearnable} and Synthetic Noise (SN)~\citep{yu2021indiscriminate} when they are transferred into unsupervised training. We generate unlearnable examples for CIFAR-10 with EMN and SN. In supervised training, we evaluate the unlearnable dataset by the test accuracy after training with CrossEntropy loss, while in unsuperivsed training, we use SimCLR~\cite{chen2020simple} to get a feature extractor and then evaluate by the accuracy of linear probing. EMN decreases the accuracy of supervised training from 93.8\% to 14.7\%, but only decreases the accuracy of unsupervised training by 0.2\%. It means that EMN has almost no unlearnable effect after being transferred into unsupervised learning and is weak in training-wise transferability. We have similar observations for SN which also lacks training-wise transferability.

\vspace{-0.05in}

\subsubsection{Data-wise Transferability}
\label{sec:pre_data_wise_transfer}

\vspace{-0.05in}

In this subsection, we investigate data-wise transferability for EMN and SN. As introduced in Eq.~\ref{EMN_formula}, EMN has been designed to induce $\boldsymbol{\delta}_i$ to have the property of minimizing the loss function in supervised training by perturbing $\boldsymbol{x}_i$. Therefore, EMN generates unlearnable examples based on target data. Next, we will show the data-wise transferability of EMN when the perturbations are transferred onto non-target data samples and non-target datasets. For $\boldsymbol{\delta}_{i}$, the target sample is $\mathbf{x}_{i}$, while the non-target samples are all the other samples $\mathbf{x}_{j}$ in $\mathcal{D}_{c}$, where $j \neq i$. For the whole perturbation set $\Delta_{\mathcal{D}_{c}}$, the non-target dataset is any other dataset $\mathcal{D}_{\tilde{c}}$. Since the perturbation $\boldsymbol{\delta}_{i}$ focuses on the target sample $\mathbf{x}_{i}$, we will first show the reduction of unlearnable effect on non-target samples. Then we will demonstrate that the perturbation set $\Delta_{\mathcal{D}_{c}}$ for $\mathcal{D}_{c}$ cannot hold the strong unlearnable effect on non-target dataset, $\mathcal{D}_{\tilde{c}}$.

\begin{figure}[t]
  \centering
  \vspace{-0.25in}
  \includegraphics[width=0.9\textwidth]{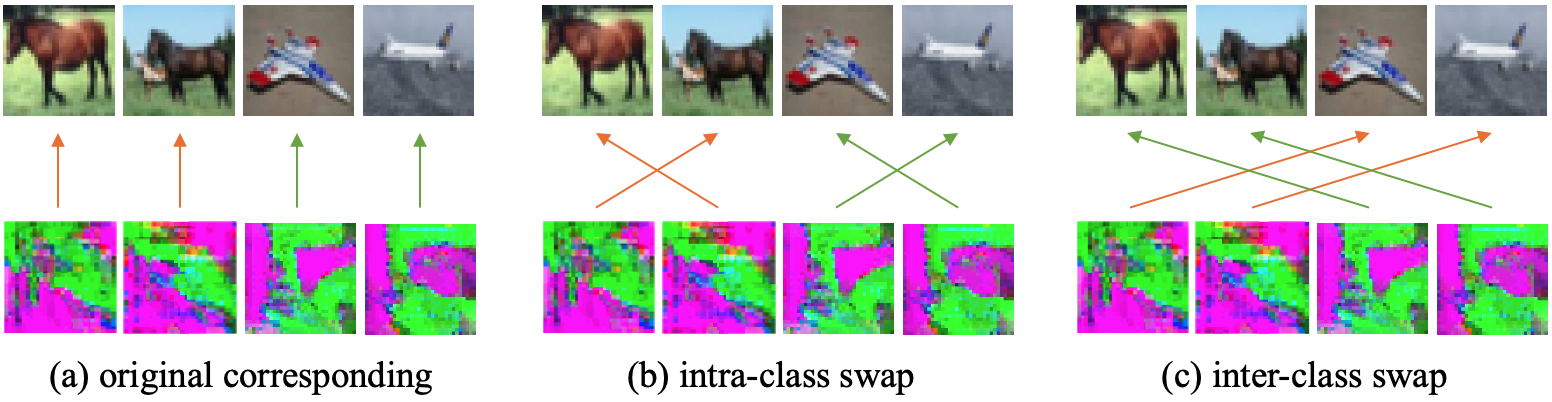}
  \vspace{-0.1in}
  \caption{Original one-to-one correspondence and two swapping methods.}
  \vspace{-0.25in}
  \label{fig:switch}
\end{figure}

\vspace{-0.05in}
On non-target samples, we change the assignment between perturbations and training samples like Fig.~\ref{fig:switch}(b) within every class or Fig.~\ref{fig:switch}(c) between the classes. Instead of perturbing $\vx_i$ with $\boldsymbol{\delta}_i$, we add $\boldsymbol{\delta}_j$ onto $\vx_i$, where $j \neq i$. 
Like Fig.~\ref{fig:switch}(b),
we construct the intra-class swapping training dataset as
$
    \mathcal{D}_{s_{\text{intra}}}=\left\{\left(\vx_{i}+\boldsymbol{\delta}_{s_{\text{intra}}(i)}, y_{i}\right)\right\}_{i=1}^{n},
$
where $s_{\text{intra}}(i)$ is the intra-class swapping function to randomly permute the correspondence between examples and perturbations within classes. For the $i$-th example, it uses the perturbation from another example in the same class:
\begin{align*}
    s_{\text{intra}}(i) = j, \text{where} \ j \neq i, y_j = y_i,
\end{align*}
Like Fig.~\ref{fig:switch}(c), we construct the inter-class swapping training dataset as 
$
    \mathcal{D}_{s_{\text{inter}}}=\left\{\left(\vx_{i}+\boldsymbol{\delta}_{s_{\text{inter}}(i)}, y_{i}\right)\right\}_{i=1}^{n}.
$
$s_{\text{inter}}$ permutes the corresponding between classes:
\begin{align*}
    s_{\text{inter}}(i) = j, \text{where} \ j \neq i, y_j \neq y_i.
\end{align*}
With $s_{\text{intra}}(i)$ and $s_{\text{inter}}(i)$, we keep the perturbation generated by EMN (i.e., $\Delta_{\mathcal{D}_{c}}$) and the clean data, (i.e., $\mathcal{D}_{c}$) unchanged, but just swap the how $\Delta_{\mathcal{D}_{c}}$ corresponds to $\mathcal{D}_{c}$. 
As shown in Table ~\ref{tab:intro_correspond}, the unlearnable effect decreases significantly under intra-class swapping and inter-class swapping.

\vspace{-0.05in}
EMN unlearnable samples are also limited in data-wise transferability on non-target datasets. We first generate the perturbations $\Delta_{\mathcal{D}_{c}}$ on CIFAR-10 and then choose another non-target dataset SVHN-small, which is downsampled from SVHN\citep{netzer2011reading}, as non-target dataset $\mathcal{D}_{\tilde{c}}$ to protect.  CIFAR-10 has ten classes, where every class has 5,000 training images. SVHN also has ten classes, but some classes have more than 5,000 training images. Thus we sample 5,000 training images in every class from SVHN and construct SVHN-small. In Table~\ref{tab:pre_transfer_dataset}, we first generate EMN on SVHN-small to protect target dataset, i.e. SVHN-small. It protects the data from unauthorized training and reduces the test accuracy to only 11.64\%. But when we use EMN generated on CIFAR-10 to protect SVHN-small, SVHN-small becomes a non-target dataset. More information is learned from SVHN-small and the test accuracy is 27.59\%, which means the non-target dataset gets less protection.





\begin{minipage}{\textwidth}
\centering
    \begin{minipage}[t]{0.48\textwidth}
    \centering
    \vspace{-0.08in}
    \makeatletter\def\@captype{table}\makeatother\caption{Comparison of supervised unlearnable examples on target and non-target samples.}
    \vspace{-0.12in}
    \label{tab:intro_correspond}
    \resizebox{\textwidth}{!}{
    \begin{tabular}{ccccc}
    \toprule
    \multirow{2}{*}{Corresponding} & \multicolumn{2}{c}{CIFAR10} & \multicolumn{2}{c}{CIFAR100} \\
    ~ & EMN & SN & EMN & SN \\
    \midrule
    Original & 15.88 & 14.07 & 6.59 & 2.13  \\
    Intra-class swap & 30.74 & 13.59 & 21.63 & 2.44 \\
    Inter-class swap & 30.15 & 13.15 & 35.50 & 2.73\\
    \bottomrule
    \end{tabular}}
    \end{minipage}
    \begin{minipage}[t]{0.48\textwidth}
    \centering
    \vspace{-0.08in}
    \makeatletter\def\@captype{table}\makeatother\caption{Comparison of supervised unlearnable examples on target and non-target datasets.}
    \vspace{-0.12in}
    \begin{tabular}{cp{1.5cm}<{\centering}p{1cm}<{\centering}}
    \toprule
    \multirow{2}{*}{Generated on} & \multicolumn{2}{c}{Tested on SVHN-small} \\
    ~ & EMN & SN \\
    \midrule
    SVHN-small & 11.64 & 8.46  \\
    CIFAR-10 & 27.59 & 9.58 \\
    \bottomrule
    \end{tabular}
    \label{tab:pre_transfer_dataset}
    \end{minipage}
\end{minipage}
\vspace{-0.1in}

We conducted similar experiments for SN~\cite{yu2021indiscriminate}. The results for non-target samples and non-target datasets are illustrated in Table~\ref{tab:intro_correspond} and Table~\ref{tab:pre_transfer_dataset}, respectively. We observe that SN is data-wise transferable. According to \cite{yu2021indiscriminate}, SN can create linear separability that the label-related imperceptible perturbation is linearly separable between different classes, as a shortcut for the optimization objective. In other words, the linear separability is only between $\boldsymbol{\delta}_i$ and $y_i$, and not related to $\vx_i$. Therefore, the SN perturbation which is totally based on linear separability can lead to the unlearnable effect and its unlearnability has good data-wise transferability. However, SN is generated by sampling from a manually designed distribution and the sampling process is incompatible with an optimization process. Thus, it is hard to directly incorporate SN into other optimization objectives to enjoy linear separability for data-wise transferability.

\vspace{-0.05in}
\subsection{Discussions}
\vspace{-0.05in}

As shown by the above preliminary studies, both EMN and SN lack training-wise transferability, and EMN has insufficient data-wise transferability. Meanwhile, 
we found that the linearly separable SN perturbation that is independent on the data has good data-wise transferability. However, the manually designed linear separability of SN is unable to be leveraged by other optimization objectives. Thus in the next section, we propose Classwise Separability Discriminant and a new framework to generate unlearnable examples with training-wise and data-wise transferability.


\vspace{-0.1in}
\section{Transferable Unlearnability from Classwise Separability Discriminant}
\label{sec:method}
\vspace{-0.05in}

In order to improve the two types of transferability of unlearnable examples, we propose the optimizable Classwise Separability Discriminant (CSD) to quantify linear separability. 
Furthermore, we propose Transferable Unlearnbale Examples (TUE) that have superior training-wise and data-wise transferability. It not only generalizes the protection scenario from supervised to unsupervised training but also maintains the unlearnability when transferred to non-target datasets.



\vspace{-0.05in}
\subsection{Classwise Separability Discriminant}
\label{sec:sepa_loss}
\vspace{-0.05in}

The linear separability lies in two factors, i.e.  intra-class distance and inter-class distance. Intuitively, better linear separability usually has smaller intra-class distance and larger inter-class distance. Because smaller intra-class distance means that the perturbations from the same class can concentrate on a small area in the input space, while larger inter-class distance means that the perturbations of different classes are far away from each other. When the overlapping area between different classes is reduced by smaller intra-class distance and larger inter-class distance, the perturbations can become features that are easily separated by even a linear classifier. For measuring the intra-class and inter-class distance, we first define the centroid of the perturbations of every class:
$
    \vc_k = \frac{1}{|\{\boldsymbol{\delta}_i:y_i=k \}|}\sum_{\{\boldsymbol{\delta}_i:y_i=k\}} \boldsymbol{\delta}_i, 
$
where $\{\boldsymbol{\delta}_i:y_i=k \}$ is the perturbations that belong to class $k$. Then we have the intra-class distance as:
\vspace{-0.05in}
\begin{align}
\small
    \boldsymbol{\sigma}_k = \frac{1}{|\{\boldsymbol{\delta}_i:y_i=k\}|}\sum_{\{\boldsymbol{\delta}_i:y_i=k\}} d(\boldsymbol{\delta}_i, \vc_k),
\end{align}

\vspace{-0.13in}
where $d(\cdot, \cdot)$ is the Euclidean distance. $\boldsymbol{\sigma}_k$ measures the average distance between the perturbation $\boldsymbol{\delta}_i$ whose label is $k$ to the centroid $\vc_k$. The inter-class distance is defined by the Euclidean distance between two centroids:
\vspace{-0.1in}
\begin{align}
    d_{i,j} = d\left(\vc_{i}, \vc_{j}\right)
\end{align}
\vspace{-0.2in}

To enhance the linear separability with an optimizable objective function, we propose Classwise Separability Discriminant (CSD):
\vspace{-0.05in}
\begin{align}
\label{eq:ls_loss}
\mathcal{L}_{\text{S}}(\{\boldsymbol{\delta}_i, y_i \}_{i=1}^n) = \frac{1}{M} \sum_{i=1}^{M} \frac{1}{M-1} \sum_{j \neq i}^{M-1} \left(\frac{\boldsymbol{\sigma}_{i}+\boldsymbol{\sigma}_{j}}{d_{i,j}}\right),
\end{align}
where $M$ is the number of classes, and $y_i$ represents the ground truth label for sample $\vx_i \in \mathcal{D}_c$. It is worth mentioning that $\mathcal{L}_{\text{S}}(\{\boldsymbol{\delta}_i, y_i \}_{i=1}^n)$ is not related to $\vx_{i}$, which means that $\boldsymbol{\delta}_{i}$ is generated independently on $\vx_{i}$.
$\boldsymbol{\sigma}_i$ and $\boldsymbol{\sigma}_j$ measure the intra-class distance of class $i$ and class $j$, while $d_{i,j}$ measures the inter-class distance between class $i$ and class $j$. The ratio between intra-class distance and inter-class distance, $\frac{\boldsymbol{\sigma}_{i}+\boldsymbol{\sigma}_{j}}{d_{i,j}}$, can show the overlapping between two classes. Better clustering distribution has smaller intra-class distance and larger inter-class distance, which lead to a lower ratio and smaller overlapping area between two classes. Therefore, lower Classwise Separability Discriminant, which indicates more compact intra-class distance and better farther inter-class distance, has better clustering effect and linear separability.
\vspace{-0.05in}
\subsection{Transferable Unlearnable Examples}
\label{TUE_methods}
\vspace{-0.05in}

As discussed in Section \ref{sec:pre_data_wise_transfer}, linear separability plays an essential role in supervised unlearnability and data-wise transferability. In order to enhance the data-wise and training-wise transferability of unlearnable examples simultaneously, we aim to incorporate linear separability into unsupervised unlearnable examples to propose Transferable Unlearnable Examples (TUE) framework. In particular, TUE uses contrastive learning as the unsupervised backbone and embeds linear separability via Classwise Separability Discriminant into unsupervised unlearnability. Most unsupervised contrastive learning algorithms, such as SimCLR \citep{chen2020simple}, MoCo \citep{chen2020improved} and SimSiam \citep{chen2021exploring}, enforce the similarity between two augmentations of the same samples such that the model can achieve representation learning. TUE generates the perturbation that not only promotes this similarity but also has linear separability by the bi-level optimization problem as follows:
\vspace{-0.05in}
\begin{align}
\label{eq:SLUCL_formula}
\min_{\theta} \underset{ \{\boldsymbol{\delta}_i:\left\|\boldsymbol{\delta}_{i}\right\|_{\infty} \leq \epsilon\}}{\min }
\sum_{i=1}^n  \mathcal{L}_{\text{CL}}\Big (f\big({\theta}, T_{1}(\vx_i  + \boldsymbol{\delta}_{i})\big), f\big({\theta}, T_{2}(\vx_i  + \boldsymbol{\delta}_{i})\big) \Big) + \lambda \mathcal{L}_{\text{S}}(\{\boldsymbol{\delta}_i, y_i \}_{i=1}^n),
\end{align}
where $\mathcal{L}_{\text{CL}}$ is the loss of contrastive learning, and $\lambda$ is the weight to balance between two loss terms. $T_{1}$ and $T_{2}$ are augmentations on input data.
The first term in Eq.~\ref{eq:SLUCL_formula} can ensure that the generated perturbation has the property of promoting the similarity between two views of augmented sample to reduce $\mathcal{L}_{\text{CL}}$. In this way, the perturbation provides easy-to-learn shortcut information for unsupervised learning such that the model will only learn to extract the perturbation instead of the intrinsic semantic information in the data. In other words, the first term is to ensure the unlearnability for unsupervised training. The second term can enhance the linear separability of the perturbations.  Furthermore, we propose to solve this bi-level optimization problem in Eq.~\ref{eq:SLUCL_formula} alternately:
\begin{small}
\begin{align}
\label{TUE_opt}
\left\{\begin{aligned}
\text{S1}:
{\theta^{(t)}} & =\underset{\theta}{\arg \min } 
\sum_{\vx_{i} \in \mathcal{D}_{c}}
    \mathcal{L}_{\text{CL}}
        \left(
            f\big({\theta}, T_{1}(\vx_i  + \boldsymbol{\delta}^{(t-1)}_{i})\big), 
            f\big({\theta}, T_{2}(\vx_i  + \boldsymbol{\delta}^{(t-1)}_{i})\big)
        \right) \\
\text{S2}:
\boldsymbol{\delta}_{i}^{(t)} & = \underset{\{\boldsymbol{\delta}_i: \left\|\boldsymbol{\delta}_{i}\right\|_{\infty} \leq \epsilon \}} {\arg \min } \mathcal{L}_{\mathrm{CL}}
\left(
    f\big({\theta}^{(t)}, T_{1}(\vx_i  + \boldsymbol{\delta}_{i})\big),
    f\big({\theta}^{(t)}, T_{2}(\vx_i  + \boldsymbol{\delta}_{i})\big)
\right)
+ \lambda \mathcal{L}_{\text{S}}(\{\boldsymbol{\delta}_i, y_i \}_{i=1}^n).
\end{aligned} \right.
\end{align}
\end{small}

\vspace{-0.1in}
In the first step (S1), we update the model parameter $\theta$ to minimize the unsupervised loss $\mathcal{L}_{\text{CL}}$, while in the second step (S2) we optimize the perturbation $\{\boldsymbol{\delta}_{i}\}$ to jointly reduce the unsupervised loss and force the linear separability among different classes. 
By the bi-level optimization on unsupervised loss and Classwise Separability Discriminant, we can generate unlearnable examples with both data-wise and training-wise transferability.

\vspace{-0.05in}
\textbf{Interpolation for data-wise transferability.}
Once we have finished the generation process on the training dataset, $\mathcal{D}_{c}$, we can use this perturbation without change on another dataset, $\mathcal{D}_{\tilde{c}}$, to make it unlearnable as well.
Nevertheless, it is possible that the number of classes in $\mathcal{D}_{c}$ or the number of samples in one class in $\mathcal{D}_{c}$ is less than $\mathcal{D}_{\tilde{c}}$. The generated perturbation may not cover every sample in $\mathcal{D}_{\tilde{c}}$. 
To solve this problem, we use interpolation to create more perturbations.
Interpolation can make use of current perturbation in $\Delta_{\mathcal{D}_{c}}$ to enlarge its size and transfer to a larger dataset. If more classes are desired, we can interpolate between two current classes in $\mathcal{D}_{c}$ to create new classes:
\begin{align}
    \vx_{k}^{*} = \alpha \vx_{i} + (1 - \alpha) \vx_{j}, \text{where} \ y_i \neq y_j.
\end{align}

\vspace{-0.1in}
If more samples in one class are required, we can interpolate within current classes in $\mathcal{D}_{c}$ to create new samples:
\vspace{-0.1in}
\begin{align}
    \vx_{k}^{*} = \alpha \vx_{i} + (1 - \alpha) \vx_{j}, \text{where} \ y_i = y_j.
\end{align}

\vspace{-0.1in}
By varying $\alpha$, more than one sample can be created from the interpolation between $\vx_{i}$ and $\vx_{j}$. Empirical results in Section~\ref{exp:transfer_between_dataset} show that this interpolation strategy works very well.



\vspace{-0.05in}
\section{Experiment}
\label{sec:experiment}
\vspace{-0.1in}


In this section, we validate the data-wise and training-wise transferability of the proposed TUE. We first introduce the experimental setups in Section~\ref{sec:setup}. In Section~\ref{sec:main_exper}, we present the experimental results on the training-wise transferability. In Section~\ref{exp:transfer_between_dataset} we demonstrate that TUE has improved data-wise transferability. 
In Section~\ref{exp:tsne_csd_dt} we illustrate the enhanced linear separability in input space by visualization.
Finally, in Section~\ref{sec:visual} we show that the two types of transferability make TUE have both classwise and samplewise characteristics as expected by case study. 

\vspace{-0.05in}
\subsection{Experimental Setups} 
\label{sec:setup}
\vspace{-0.05in}


\textbf{Datasets.} The datasets include CIFAR-10 and CIFAR-100~\citep{krizhevsky2009learning}, which contain 50,000 training images and 10,000 test images, and SVHN~\citep{netzer2011reading}, which contains 73,257 training images of ten classes and 26,032 test images. 
We randomly sample from SVHN to construct SVHN-small where the number of training images in every class is no more than 5000. 
\vspace{-0.05in}

\textbf{Generation process and Evaluation settings.} We compare our method with representative unlearnable methods in both supervised and unsupervised training.
All the perturbations are generated by PGD~\citep{madry2018towards} on ResNet-18 and constrained by $\left\|\delta_{i}\right\|_{\infty} \leq \epsilon$ where $\epsilon = 8/255$. We use CrossEntropy as the objective function in supervised training. 
and linear probing after contrastive pre-training in unsupervised training. The supervised model is trained for 200 epochs. The unsupervised model is pre-trained for 1000 epochs and then fine-tuned for 100 epochs. 
More hyperparamaters for generation process and evaluation can be found in Appendix \ref{append_settings}.
\vspace{-0.05in}

\textbf{Baselines.} 
We use three representative unlearnable examples as baselines, which are Error-Minimizing Noise (EMN)~\citep{huang2020unlearnable}, Unlearnable Contrastive Learning (UCL)~\citep{he2022indiscriminate} and Synthetic Noise (SN)~\citep{he2022indiscriminate}.  

\vspace{-0.05in}
\textbf{Backbones.} We tested TUE on three different unsupervised backbones, SimCLR~\citep{chen2020simple}, MoCo~\citep{chen2020improved} and SimSiam~\citep{chen2021exploring}. The backbones are also used in UCL.
\vspace{-0.05in}


\begin{table}[h]
  \caption{\small Performance of unlearnable effects on different algorithms and datasets in supervised and unsupervised training. This table reports the accuracy (\%) of supervised training and the accuracy (\%) of linear probing after contrastive pre-training (SimCLR, MoCo, and SimSiam). 
  }
  \label{main_result}
  \centering
  \resizebox{.7\textwidth}{!}{
  \begin{threeparttable}
  \begin{tabular}{c|cccc}
    \toprule
    {Dataset} & \multicolumn{2}{c}{CIFAR-10} & \multicolumn{2}{c}{CIFAR-100}            \\
    \midrule
    {SimCLR}  & Supervised & Unsupervised & Supervised & Unsupervised \\
    \midrule
    {Clean Data} & 93.79 & 90.04 & 74.49 & 63.68     \\
    \midrule
    EMN & {14.74} & 89.79 & {5.23} & 62.00       \\
    SN & {19.23} & 88.93 & {2.13} & 62.31  \\
    UCL & 92.86 & \textbf{47.78} & 72.17 & \textbf{16.68} \\
    TUE & \textbf{10.67} & {52.38} & \textbf{0.76} & {19.51} \\
    
    \midrule
    {MoCo}  & Supervised & Unsupervised & Supervised & Unsupervised \\
    \midrule
    {Clean Data} & 93.79 & 89.90 & 74.49 & 63.03 \\
    \midrule
    EMN & {14.74} & 89.18 & {5.23} &  60.62 \\
    SN & {19.23} & 89.32 & {2.13} &  61.81 \\
    UCL & 92.62 & \textbf{44.24} & 71.59 & \textbf{18.74} \\
    TUE & \textbf{10.06} & 63.38 & \textbf{1.09} & 23.6 \\
    
    \midrule
    {SimSiam}  & Supervised & Unsupervised & Supervised & Unsupervised \\
    \midrule
    {Clean Data} & 93.79 & 90.59 & 74.49 & 64.69     \\
    \midrule
    EMN& {14.74} & 91.43 & {5.23} &  65.96      \\
    SN & {19.23} & 91.54 & {2.13} & 66.83  \\
    UCL & 93.50 & \textbf{30.43} & 71.84 & \textbf{4.64} \\
    TUE & \textbf{10.03} & 35.57 & \textbf{1.21} & 6.17  \\
    
    \bottomrule
  \end{tabular}
  \end{threeparttable}
  }
  \vspace{-0.2in}
\end{table}

\vspace{-0.07in}
\subsection{Training-wise transferable unlearnability}
\label{sec:main_exper}
\vspace{-0.07in}

As mentioned above, existing unlearnable methods focus on one training setting, either supervised or unsupervised training. In this work, we use the proposed TUE to make the dataset unlearnable in both two training settings and thus protect the data from unauthorized training in a comprehensive way. 
In Table~\ref{main_result}, we reported the comparison on CIFAR-10 and CIFAR-100 with supervised training by CrossEntropy Loss and three unsupervised training method, SimCLR~\citep{chen2020simple}, MoCo~\citep{chen2020improved} and SimSiam~\citep{chen2021exploring}. The unlearnable examples are evaluated by two training settings, supervised and unsupervised training. We use TUE to embed the linear separability into SimCLR, MoCo and SimSiam, respectively, and get three protected datasets. When testing with unsupervised training, we use the corresponding unsupervised algorithm that is used to generate TUE.
\vspace{-0.05in}

Table~\ref{main_result} shows the model accuracy on clean test data after being trained on unlearnable data. 
First, we observe that previous unlearnability methods can only work under one training setting, and have almost no protection under the other training setting. Specifically, EMN and SN reduces the test accuracy of supervised training to less than 20\% on CIFAR-10 and less than 6\% on CIFAR-100 but cannot transfer well in all of the three unsupervised trainings. UCL reduces the accuracy of unsupervised training to less than 50\% on CIFAR-10 and less than 20\% on CIFAR-100 but has almost no protection from supervised training. Second, TUE can hold the unlearnability in both supervised and unsupervised training settings. It reduces the supervised accuracy to around 10\% on CIFAR-10 and 1\% on CIFAR-100 and reduces the unsupervised accuracy to a comparable level to UCL.
In particular, it can even perform better than EMN in supervised training. The test accuracy of unauthorized classifiers trained on TUE is around 4\% lower than EMN on both CIFAR-10 and CIFAR-100. Because the enhanced linear separability causes that supervised model focuses more on TUE perturbation than EMN perturbation. In summary, from Table~\ref{main_result}, we can see that TUE has good training-wise transferability, which can maintain the unlearnable effect when being transferred from supervised training to unsupervised training and it is the only one that maintains the unlearnable effects under two training settings.

\vspace{-0.05in}
\subsection{Data-wise transferable unlearnability}
\label{exp:transfer_between_dataset}
\vspace{-0.05in}

After demonstrating the training-wise transferability of TUE, in this subsection, we show that TUE also has better data-wise transferability, i.e., the perturbations generated with TUE on one dataset can also protect any other datasets from unauthorized supervised training. Following the experimental settings in the preliminary studies of Section~\ref{sec:pre_data_wise_transfer}, we test the data-wise transferability first on non-target data samples, and then on non-target datasets. The results are reported in Table~\ref{tab:transfer_sample} and Table~\ref{tab:transfer_dataset}, respectively.


\vspace{-0.08in}
\begin{minipage}{\textwidth}
    \begin{minipage}[t]{0.5\textwidth}
    \centering
    \makeatletter\def\@captype{table}\makeatother\caption{\small Test accuracy (\%) with different correpondings between train data and perturbations.}
    \vspace{-0.12in}
    \label{tab:transfer_sample}
    \resizebox{.98\textwidth}{!}{
    \begin{tabular}{ccccc}
    \toprule
    Dataset & Methods & Original & Intra & Inter \\
    \midrule
    CIFAR-10 & EMN & 15.88 & 30.74 & 33.91 \\
    & SN & 14.07 & 13.59 & 13.15 \\
    & TUE (SimCLR) & 10.67 & 10.16 & 10.90 \\
    & TUE (MoCo) & 10.06 & 12.04 & 8.57 \\
    & TUE (SimSiam) & 10.03 & 10.25 & 10.47 \\
    \midrule
    CIFAR-100 & EMN & 6.59 & 21.63 & 35.50 \\
    & SN & 2.13 & 2.44 & 2.73 \\
    & TUE (SimCLR) & 0.76 & 1.11 & 1.13 \\
    & TUE (MoCo) & 1.09 & 1.25 & 3.63 \\
    & TUE (SimSiam) & 1.21 & 1.28 & 1.08 \\
    \bottomrule
    \end{tabular}}
    \end{minipage}
    \begin{minipage}[t]{0.5\textwidth}
    \centering
    \makeatletter\def\@captype{table}\makeatother\caption{\small Test accuracy (\%) of transferring the perturbation generated on CIFAR-10 to different datasets.}
    \vspace{-0.12in}
    \resizebox{.94\textwidth}{!}{
    \begin{tabular}{cccc}
    \toprule
    & {SVHN-small} & {CIFAR-100} & {SVHN} \\
    \midrule
    EMN & 27.59 & 21.80 & 24.72 \\
    SN & 9.58 & 9.35 & 7.77 \\
    TUE (SimCLR) & 9.77 & 10.53 & 11.72 \\
    TUE (MoCo) & 11.29 & 8.32 & 13.95 \\
    TUE (SimSiam) & 10.28 & 5.10 & 12.93 \\
    \bottomrule
    \end{tabular}}
    \label{tab:transfer_dataset}
    \end{minipage}
\end{minipage}


First, for non-target data samples, we compare the performance under the original correspondence, intra-class swapping and inter-class swapping between samples and perturbations in the same dataset. In Table~\ref{tab:transfer_sample}, TUE can maintain similar test accuracy under different correspondences on CIFAR-10 and CIFAR-100. For example, on CIFAR-100, the test accuracy of EMN in intra-class and inter-class swapping is around, respectively, 15\% and 29\% higher than that in the original correspondence, but for TUE, the difference in the test accuracy under different correspondences is less than 3\%. Although we can also notice this data-wise transferability in SN, it is not an optimizable method and is unable to be used for improving the training-wise transferability.

\vspace{-0.05in}
Second, to verify that the proposed TUE can maintain unlearnability when transferred to different datasets, we generate TUE on CIFAR-10 and test the performance of transferring onto three datasets, SVHN-small, CIFAR-100 and SVHN. TUE on CIFAR-10 can be directly transferred to SVHN-small without interpolation, while CIFAR-100 requires interpolation for more classes and SVHN requires interpolation for more samples. The baseline methods are also interpolated for comparison.
In Table~\ref{tab:transfer_dataset}, we can observe that on all the datasets, TUE can maintain the unlearnable effect after being transferred onto all the other datasets. 
The enhanced linear separability of TUE ensures that the test accuracy is low. For example, EMN has poorer protection after transferred onto SVHN-small with the accuracy of 27.59\%, but TUE can maintain the unlearnability on non-target datasets for all the backbones at around 10\% accuracy.
\vspace{-0.05in}

From Table~\ref{tab:transfer_sample} and Table~\ref{tab:transfer_dataset}, we can draw the conclusion that the proposed TUE has better data-wise transferability than optimization-based unlearnable methods like EMN, and our strategy to transfer TUE to protect other datasets can be very efficient. 


 
\vspace{-0.07in}
\subsection{Linear Separability of Transferable Unlearnable Examples}
\label{exp:tsne_csd_dt}
\vspace{-0.07in}


In this subsection, we aim to better understand how linear separability influences unlearnable examples by visualization.
To better visualize linear separability of the perturbations, we use t-SNE to show the perturbation in the input space. Although t-SNE cannot accurately describe the high-dimension space, it is useful to observe the separability of the perturbations in the input space. We compare the perturbations in the input space in the following three groups. 

\begin{figure}[th]
  \centering
  \vspace{-0.15in}
  \includegraphics[width=0.8\textwidth]{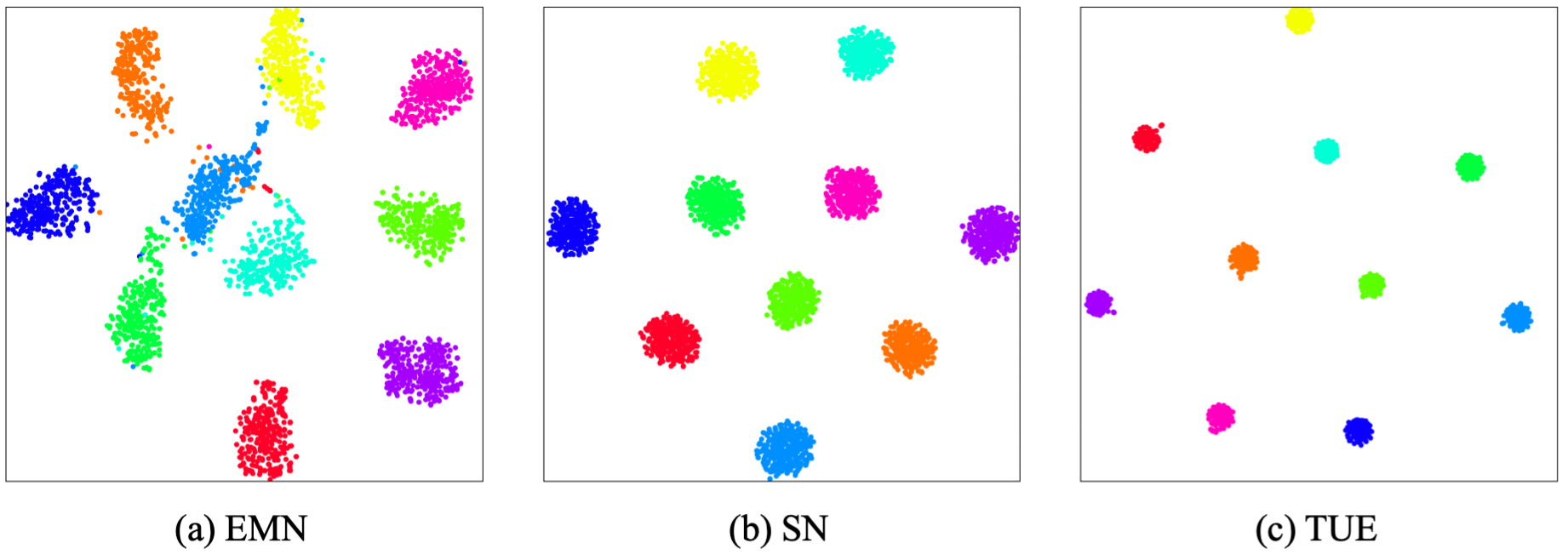}
  \vspace{-0.15in}
  \caption{t-SNE visualization of linear separability of supervised unlearnable examples.}
  \vspace{-0.1in}
  \label{linear_tsne1}
\end{figure}


\vspace{-0.05in}
\textbf{Comparison of Linear separability in supervised unlearnable examples.} In Fig.~\ref{linear_tsne1}, we show the perturbations of EMN, SN and TUE to understand why linear separability is more effective than EMN when used in supervised classification. In Fig.~\ref{linear_tsne1}, the perturbations of different classes in TUE and SN are clearly separated and have no overlapping. In contrast, EMN has some data points mixed with other classes which are confusing when used in classification. Meanwhile the clusters in EMN do not contract as well as TUE and SN. Since better linear separability can provide useful information about classification, the supervised model will focus on TUE and SN more than EMN and ignore the semantic features in clean images.

\begin{figure}[h]
  \centering
  \includegraphics[width=0.53\textwidth]{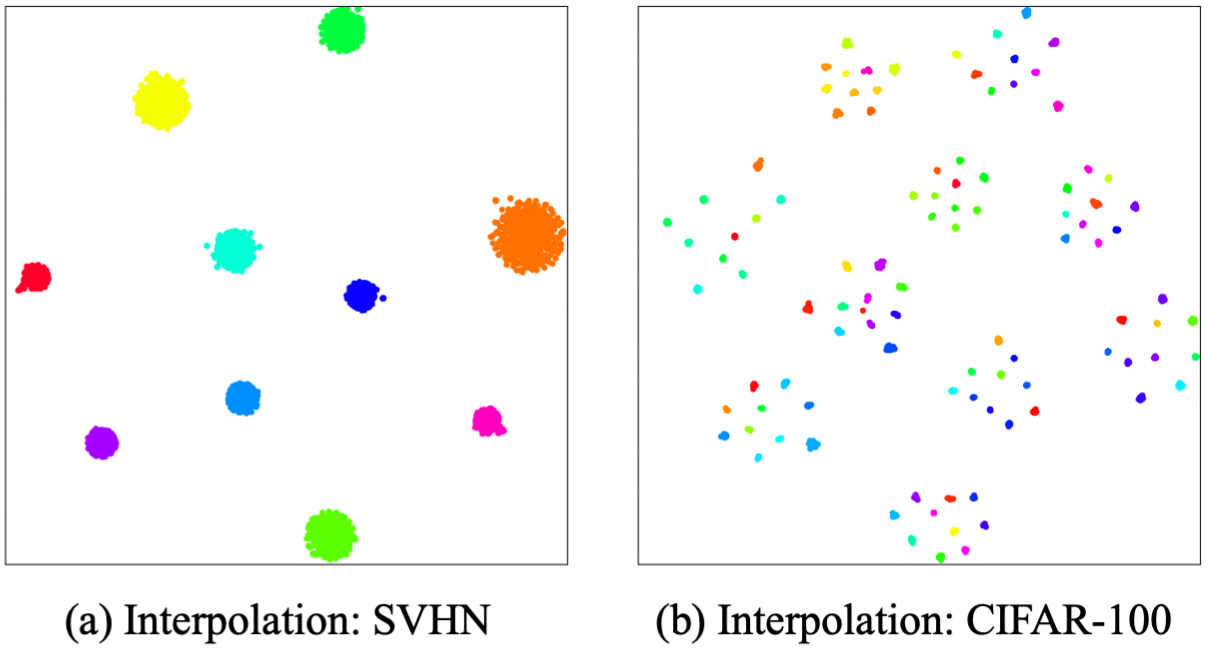}
  \vspace{-0.15in}
  \caption{t-SNE visualization of linear separability of two examples of interpolation.}
  \vspace{-0.1in}
  \label{linear_tsne2}
\end{figure}


\textbf{Linear separability in Interpolation.} In  Fig.~\ref{linear_tsne2}, we show that the interpolation on TUE from CIFAR-10 to SVHN and CIFAR-100 can also hold linear separability which makes it data-wise transferable. 
We can observe that in the interpolation for SVHN, although the size of each class is different, they can still hold clear linear separability and unlearnbility. In SVHN, the interpolation within classes creates more samples for one class. 
Since the numbers of examples in different classes in SVHN are different, it can be observed that some clusters are much larger than others. 
For CIFAR-100, more new classes are created and they still keep good linear separability. So the supervised model trained on interpolation-based unlearnable SVHN and CIFAR-100 will learn nothing but the perturbations and perform poorly on clean test data which has no perturbations.

\begin{figure}[h]
  \centering
  \vspace{-0.15in}
  \includegraphics[width=0.53\textwidth]{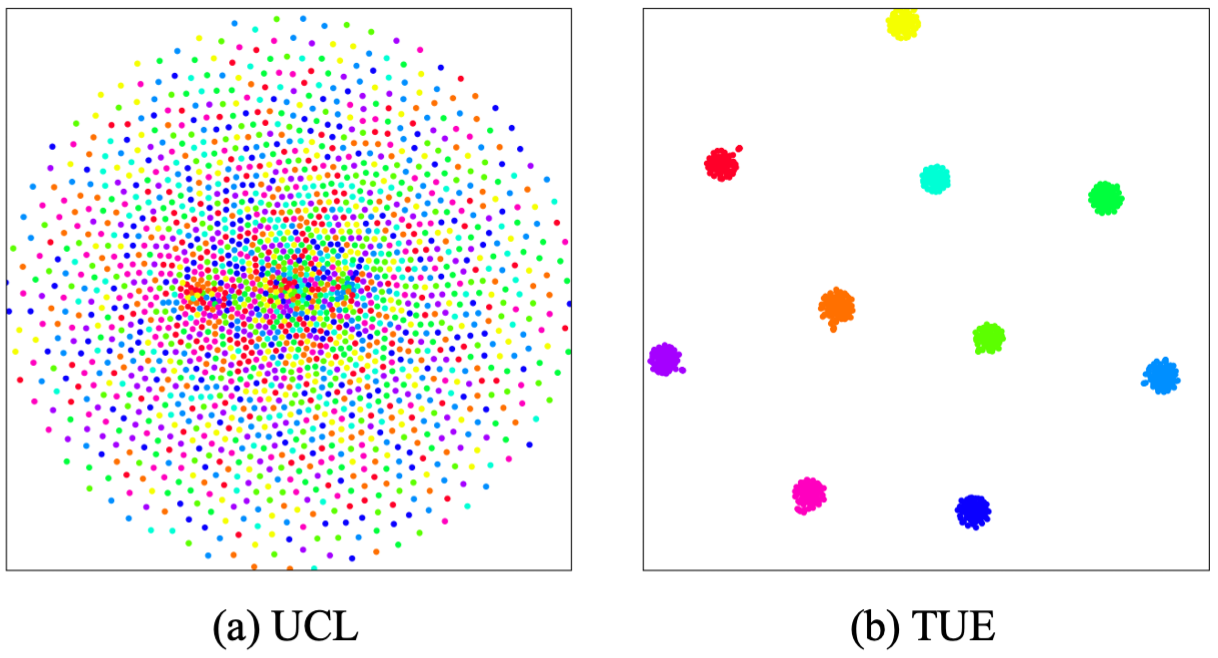}
  \vspace{-0.15in}
  \caption{t-SNE visualization of linear separability of supervised and unsupervised unlearnability.}
  \vspace{-0.1in}
  \label{linear_tsne3}
\end{figure}


\vspace{-0.05in}
\textbf{Comparison of Linear separability between supervised and unsupervised unlearnability.} In  Fig.~\ref{linear_tsne3}, we show the perturbations of UCL and TUE in input space to understand why UCL cannot help with unlearnable effects in supervised training. TUE is linear separable, which can be used to protect the data from both supervised and unsupervised training. But for UCL, all the perturbations are mixed, they cannot provide the class information for supervised training. So for a supervised model, UCL is not a helpful feature and will not vanquish the semantic feature.

\vspace{-0.05in}
\subsection{Case Study}
\label{sec:visual}
\vspace{-0.05in}

\begin{figure}[h]
  \centering
  \vspace{-0.1in}
  \includegraphics[width=1\textwidth]{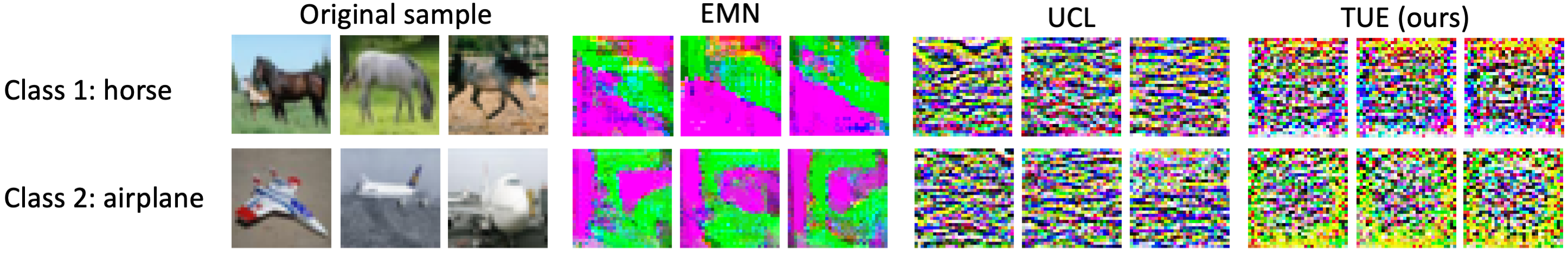}
  \vspace{-0.25in}
  \caption{Visualization of the unlearnable examples of EMN, UCL and TUE(ours). We rescale the perturbation from [-8/255, 8/255] to [0,1] for better visualization.}
  \vspace{-0.1in}
  \label{case_study_pic222}
\end{figure}

\vspace{-0.05in}
In this subsection, by showing the patterns of the perturbations generated by different methods, we can deepen our understanding on  linear separability in TUE and observe the classwise and samplewise characteristics, which reflect the basic idea behind training-wise transferability.

\vspace{-0.05in}

We illustrate the samples and unlearnable perturbations from two classes of CIFAR-10, i.e., horse and airplane, in Figure~\ref{case_study_pic222}. First, we find that EMN and TUE both have classwise noise patterns. Every noisy sample has similar features in one class, while every class has a different feature pattern from other classes. This is a new perspective to show that TUE has linear separability and perturbations from the same class are clustered in input space.
Second, in UCL, the perturbations can be more diverse and related to each sample, which indicates that unsupervised unlearnability requires more complex feature patterns. We cannot see the classwise patterns like EMN. Because UCL uses no label information and has no concept of classes.
Third, we notice that TUE has both classwise and samplewise characteristics as expected. The classwise feature can be the easy-to-learn feature for the supervised model, while the samplewise feature can be effective in reducing unsupervised loss, which makes the unlearnability transferable from supervised training to unsupervised training.
In summary, from the case study, we can observe that the co-existing classwise and samplewise features provide supervised and unsupervised protection at the same time. The linear separability makes the supervised unlearnability data-wise transferable.


\vspace{-0.05in}
\section{Conclusion}
\vspace{-0.1in}
    
We reveal the limitation of training-wise and data-wise transferability in existing unlearnable examples and propose Classwise Separability Discriminant to enhance transferability. The proposed Transferable Unlearnable Examples (TUE) can be transferred from supervised to unsupervised training with the unlearnable effect maintained and can protect data from unauthorized usage in a comprehensive way. Meanwhile, the unlearnability of TUE is transferable across datasets. It can generate unlearnbility once for all, which makes unlearnable examples more practical and efficient. TUE greatly pushes the boundary of existing unlearnable methods that can only work on the target data and target training setting. We improve the data-wise and training-wise transferability of unlearnable examples and provide more flexible, practical, and comprehensive protections from unauthorized data usage.

\bibliography{iclr2023_conference}
\bibliographystyle{iclr2023_conference}

\appendix

\section{Existing unlearnable methods cannot protect unlabeled dataset in a training-wise trasnferable way}
\label{append_unsupervised}
\begin{figure}[h]
  \centering
  \includegraphics[width=0.5\textwidth]{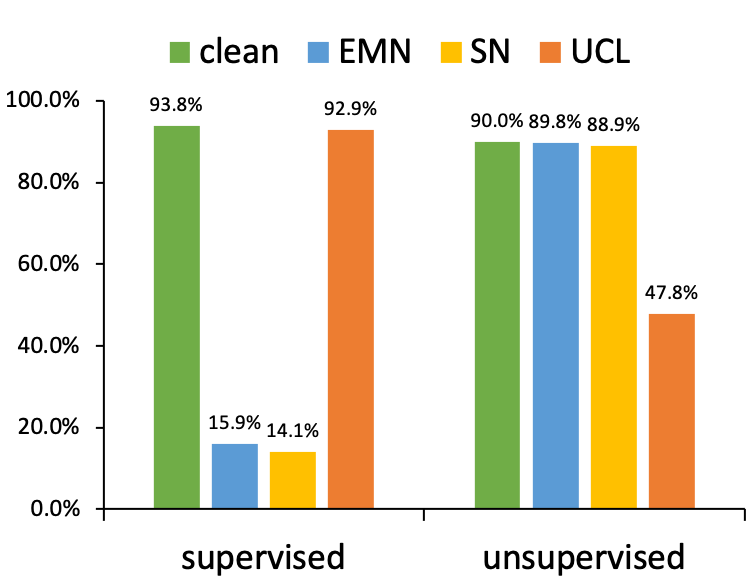}
  \caption{Accuracy on test data of classifiers produced by supervised training and unsupervised training on the unlearnable examples from CIFAR-10}
  \label{pre_trainingwise_unsupervised}
\end{figure}

Previous unlearnable methods can only protect data from one unauthorized training setting. According to whether the dataset is annotated, unlearnable examples can be classified in two settings, supervised and unsupervised unlearnable examples, as mentioned in Section \ref{sec:relared}. The supervised unlearnable methods, like EMN, can create unlearability based on the label information and protect the data from supervised training, but these methods cannot prevent the unauthorized parties from training the data with an unsupervised algorithm. Similarly, the unsupervised unlearnable examples produced by Unleanrable Contrastive Learning (UCL)~\cite{he2022indiscriminate} can destroy the unsupervised training but they are still learnable in the supervised training. In Fig. \ref{pre_trainingwise_unsupervised}, EMN decreases the accuracy of supervised training from 93.8\% to 14.7\%, but only decreases the accuracy of unsupervised training by 0.2\%. Similarly, SN decreases supervised training to 14.1\%, but only decreases unsupervised training by 1.1\%. Supervised unlearnable examples have almost no training-wise transferability to unsupervised training. UCL, which produces unsupervised unlearnable examples, can protect data from unsupervised training, but cannot be transferred well to supervised training. Under unsupervised training, UCL can decrease the test accuracy from 90.0\% to 47.8\%, but under supervised training, it can only decrease the accuracy by 0.9\%.

\section{Details of experimental settings}
\label{append_settings}

\subsection{Settings of generation process}

All the baselines and our proposed TUE are generated in the way of PGD \cite{madry2018towards}, except for SN, which is sampled from a manually designed distribution following the algorithm in \cite{yu2021indiscriminate}. For EMN, UCL, and TUE, the generation process is an alternate optimization between model parameters and perturbations. For example, TUE is optimized in the way of Eq. \ref{TUE_opt}. In EMN, after every epoch of optimization on model parameters, the whole set perturbations are optimized for one epoch by PGD-20. In TUE, different backbones are set to different schedule. For TUE (SimCLR), the model parameters are trained for 40 epochs, and after every 1/5 epoch of optimization on model parameters, the whole set perturbations are optimized for one epoch by PGD-20.
For TUE (MoCo), the model parameters are trained for 200 epochs, and after every epoch of optimization on model parameters, the whole set perturbations are optimized for one epoch by PGD-5.
For TUE (SimSiam), the model parameters are trained for 50 epochs, and after every 1/4 epoch of optimization on model parameters, the whole set perturbations are optimized for one epoch by PGD-20. UCL with SimCLR and SimSiam have the same settings as TUE. UCL with MoCo uses PGD-10 and the other settings in the schedule is the same as TUE (MoCo). Finally, all the perturbations are constrained by $l_\infty$ norm, i.e. $\left\|\delta_{i}\right\|_{\infty} \leq \epsilon$ where $\epsilon = 8/255$.

\subsection{Settings of evaluation stage}

To evaluate the unlearnable effect, we use CrossEntropy as the objective function in the unauthorized supervised training. We use linear probing after contrastive pre-training to evaluate the unlearnable effect in unsupervised training. The supervised model is trained for 200 epochs. The unsupervised model is pre-trained for 1000 epochs by unsupervised contrastive learning and then fine-tuned for 100 epochs by linear probing. The details for the hyperparameters are listed in Table \ref{tab:detailed_setting}.

\begin{table}[t]
  \caption{Hyperparameters of evaluation stage}
  \label{tab:detailed_setting}
  \centering
  \resizebox{\textwidth}{!}{
  \begin{tabular}{c|ccccccc}
    \toprule
    \multirow{2}{*}{Hyperparameter} & \multirow{2}{*}{Supervised} &  \multicolumn{3}{c}{Unsupervised pretraining} &  \multicolumn{3}{c}{Linear probing} \\
    ~ & ~ & SimCLR & MoCo & SimSiam & SimCLR & MoCo & SimSiam \\
    \midrule
    Epoch & 200 & 1000 & 1000 & 1000 & 100 & 100 & 100\\
    Optimizer & SGD & SGD & Adam & SGD & Adam & SGD & SGD\\
    Learning Rate & 0.1 & 0.06 & 0.3 & 0.06 & 0.001 & 30 & 30\\
    LR Scheduler & CosineAnnealingLR (CA) & - & CA & CA & - & CA & CA \\
    Encoder Momentum & - & - & 0.99 & - & - & - & - \\
    Loss Function & CrossEntropy (CE) & InfoNCE & InfoNCE & \makecell{Similarity between \\ positive samples} & CE & CE & CE\\
    \bottomrule
  \end{tabular}}
\end{table}

\end{document}